\documentclass[sn-mathphys,Numbered]{sn-jnl}

\usepackage{graphicx}%
\usepackage{multirow}%
\usepackage{amsmath,amssymb,amsfonts}%
\usepackage{amsthm}%
\usepackage{mathrsfs}%
\usepackage[title]{appendix}%
\usepackage{xcolor}%
\usepackage{ulem}
\usepackage{textcomp}%
\usepackage{manyfoot}%
\usepackage{booktabs}%
\usepackage{algorithm}%
\usepackage{algorithmicx}%
\usepackage{algpseudocode}%
\usepackage{listings}%
\usepackage{graphicx}
\usepackage{outlines}

\theoremstyle{thmstyleone}%

\theoremstyle{thmstyletwo}%

\theoremstyle{thmstylethree}%

\definecolor{mygreen}{rgb}{0.0, 0.65, 0.0}

\begin{document}


\title[Towards Controllable Video Synthesis of Routine and Rare OR Events]{Towards Controllable Video Synthesis of Routine and Rare OR Events}




\author[1,2]{\fnm{Dominik} \sur{Schneider}}
\equalcont{Co-first author.}
\author[1]{\fnm{Lalithkumar} \sur{Seenivasan}} 
\equalcont{Co-first author.}
\author[1]{\fnm{Sampath} \sur{Rapuri}}
\author[1]{\fnm{Vishalroshan} \sur{Anil}}
\author[1]{\fnm{Aiza} \sur{Maksutova}}
\author[1]{\fnm{Yiqing} \sur{Shen}}
\author[1]{\fnm{Jan Emily} \sur{Mangulabnan}}
\author[1]{\fnm{Hao} \sur{Ding}}
\author[1,3]{\fnm{Jose L.} \sur{Porras}}
\author[3]{\fnm{Masaru} \sur{Ishii}}
\author*[1]{\fnm{Mathias} \sur{Unberath}} \email{unberath@jhu.edu}

\affil*[1]{\orgname{Johns Hopkins University}, \orgaddress{\city{Baltimore}, \postcode{21211}, \state{MD}, \country{USA}}}
\affil*[2]{\orgname{Technical University Munich}, \orgaddress{\city{Munich}, \postcode{80333}, \state{BY}, \country{Germany}}}
\affil[3]{\orgname{Johns Hopkins Medical Institutions}, \orgaddress{\city{Baltimore}, \postcode{21287}, \state{MD}, \country{USA}}}

%
%

\abstract{
\textbf{Purpose:}
Curating large-scale datasets of operating room (OR) workflow, encompassing rare, safety-critical, or atypical events, remains operationally and ethically challenging. This data bottleneck complicates the development of ambient intelligence for detecting, understanding, and mitigating rare or safety-critical events in the OR. \\
\textbf{Methods:} This work presents an OR video diffusion framework that enables controlled synthesis of rare and safety-critical events. The framework integrates a geometric abstraction module, a conditioning module, and a fine-tuned diffusion model to first transform OR scenes into abstract geometric representations, then condition the synthesis process, and finally generate realistic OR event videos. Using this framework, we also curate a synthetic dataset to train and validate AI models for detecting near-misses of sterile-field violations.\\
\textbf{Results:} In synthesizing routine OR events, our method outperforms off‑the‑shelf video diffusion baselines, achieving lower FVD/LPIPS and higher SSIM/PSNR in both in‑ and out‑of‑domain datasets. Through qualitative results, we illustrate its ability for controlled video synthesis of counterfactual events. An AI model trained and validated on the generated synthetic data achieved a RECALL of $70.13$\% in detecting near safety-critical events. Finally, we conduct an ablation study to quantify performance gains from key design choices. \\

\textbf{Conclusion:} Our solution enables controlled synthesis of routine and rare OR events from abstract geometric representations. Beyond demonstrating its capability to generate rare and safety-critical scenarios, we show its potential to support the development of ambient intelligence models.}



\maketitle

\section{Introduction}\label{sec:introduction}
With the operating room (OR) being central to patient care~\cite{saeedian2019operating} and hospital economics~\cite{healey2015improving}, driving efforts toward OR ambient intelligence can enhance hospital performance both clinically and financially.  Clinically, enabling automated OR workflow analysis and optimization reduces intraoperative risk and improves patient outcome: automated detection of safety‑critical events (e.g., sterile‑field breaches); shorter anesthesia exposure benefits patients~\cite{phan2017anesthesia}; each additional 60 minutes of surgery increases the odds of surgical‑site infection by 37\%~\cite{cheng2017prolonged}; improved coordination is associated with fewer complications~\cite{koch2020associations}; and shorter case duration improves OR throughput, increasing access to more patients in need of care. Financially, global surgical demand exceeds existing hospital capacity~\cite{meara2015global, weiser2008estimation}. With constant demand for surgery and ORs generating roughly 60-70\% of hospital revenue~\cite{healey2015improving} and accounting for 35-40\% of hospital expenses~\cite{healey2015improving}, enabling automated OR workflow optimization could increase surgical throughput, increasing hospital revenue. Furthermore, optimizing OR workflow could potentially allow for better resource utilization and operational efficiency~\cite{vladu2024enhancing}, reducing hospital costs. Although emerging AI solutions bring OR ambient intelligence within reach, progress is constrained by the lack of comprehensive datasets that capture the full spectrum of OR events necessary for model development.

Curating rare, safety-critical, or atypical  OR events at scale is operationally and ethically challenging. In clinical practice, generating that data is practically difficult due to privacy and access constraints, site and procedure variability, and the inherent rarity of critical events. Deliberately eliciting safety‑critical rare events -- sterile field violations, equipment handoffs, or deviations from standard protocols -- for enriching the dataset is ethically impermissible and risks patient harm. Manual curation and staged reenactments are not scalable, given the breadth of clinical variability, staffing limitations, and operational disruption. There is a clear need for scalable methods to generate OR events on demand -- with rich procedural variation and rare, safety‑critical scenarios -- to support the development of OR ambient intelligence.

Advancing controllable and scalable data curation methods, we introduce an OR video‑diffusion framework conditioned on an abstract geometric scene representation to curate synthetic videos of routine and rare OR events. The framework represents the OR workspace and events using simple geometric primitives: ellipsoids for personnel, the patient, and equipment. Given an initial OR scene and abstract geometric representations of the intended OR event as conditioning input, it generates synthetic videos of the specified events. The event conditioning is modeled either from prior routines derived from known OR events or from user‑defined trajectories on the abstract geometric representations. Our key contributions are:

\begin{itemize}
    \item We introduce an abstract, geometry conditioned OR diffusion framework with a novel geometric abstraction and conditioning module, enabling controlled, scalable synthesis of OR‑event videos via ellipsoid‑based entity representation and trajectory sketches.
    
    \item We demonstrate the video synthesis of routine, rare, atypical, and safety-critical OR events (sterile-field violations) that would otherwise be difficult to collect due to practical/ethical reasons. 
    
    \item We curate a synthetic dataset and train AI models to detect sterile‑field violations, achieving 70\% recall (sensitivity) and demonstrating the framework’s potential to enable scalable data curation for ambient intelligence development.
    
    \item Additionally, we augment the baseline fine-tuning with a PatchGAN loss to improve the local realism and fidelity of the synthesized videos. 

\end{itemize}


\section{Related Work}\label{sec:relatedwork}

Generative models for general-purpose video synthesis have been widely explored from early GAN-based \cite{Saito2016-lj} approaches to more modern diffusion model frameworks \cite{Peebles2022-kv}. Among the general-purpose video diffusion models, Stable Video Diffusion (SVD) \cite{blattmann2023stable} is a popular image-to-video diffusion model, incorporating an image frame and an optional text prompt as conditioning inputs. In contrast, the WAN family of models \cite{wan2025wan} represents a series of text-conditioned diffusion approaches. While these models achieve strong performance, they utilize natural-language prompts and/or single keyframes as conditioning inputs, which lack fine-grained control over object positioning, orientation, and interactions. In the surgical domain, generative models have been adopted for simulation \cite{Chen2025-dh}, using controllable conditioning inputs to guide the video generation process. Typically, these conditioning signals are class labels, text prompts, reference images or videos, or trajectory information, essential for accurately modeling complex and high-risk surgical workflows.
%
To our knowledge, no prior work has achieved controllable generation of the ambient operating room environment, including staff movement, equipment repositioning, and safety-critical events. 


\section{Method}
\begin{figure}[t]
    \centering
    \includegraphics[width=0.95\linewidth]{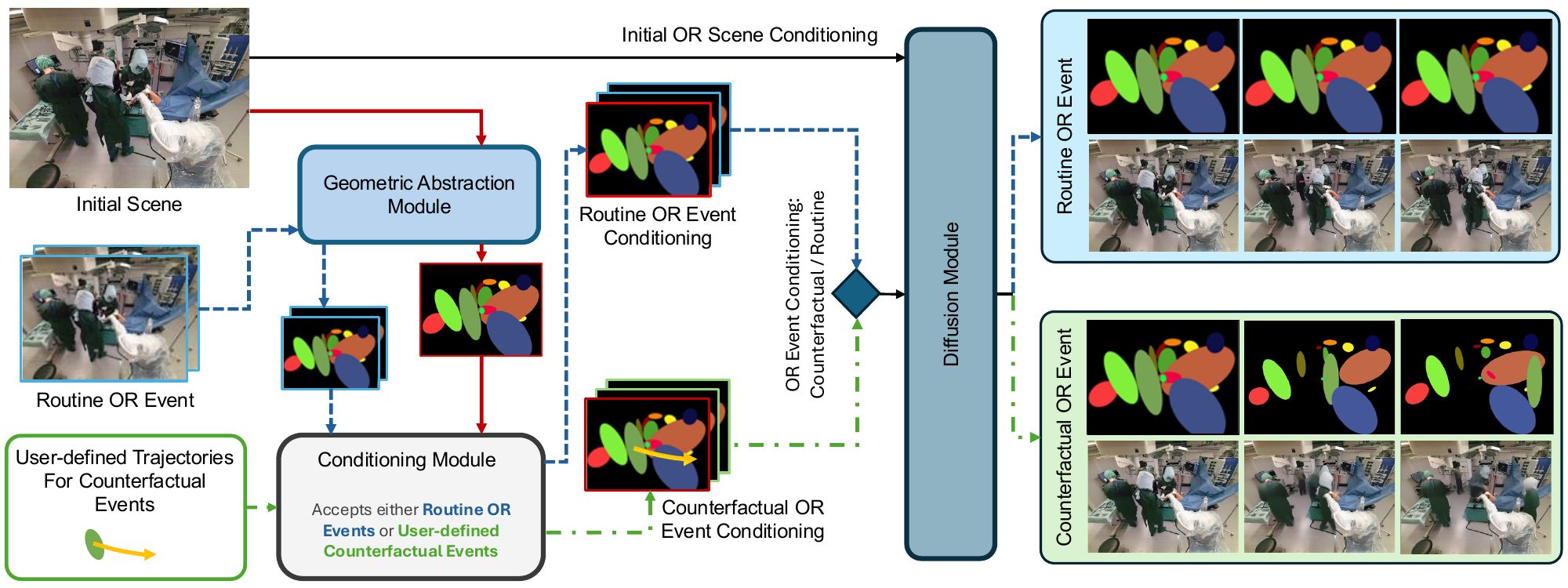}
\caption{abstract, geometry conditioned OR diffusion framework consists of three main modules: \uline{(i) Geometric Abstraction Module} converts the initial OR scene into an abstract geometric scene representation using ellipsoids. \uline{(ii) Conditioning Module} generates temporal sequences of abstract geometric scenes through two pathways: from routine OR events (blue dash path), or from incorporating user-defined trajectories (dashdotted green path). \uline{(iii) Diffusion Module} synthesizes videos of OR events conditioned on the initial scene and the geometric sequences.}
\label{fig:method_overview}
\end{figure}

Our proposed abstract, geometry conditioned OR diffusion framework reformulates a video-to-video diffusion task into a controlled OR event video generation task conditioned via abstract geometric scene representations. The framework incorporates three main modules: (i) geometric abstraction module, (ii) conditioning module, and (iii) diffusion module (Fig.~\ref{fig:method_overview}). Given an initial OR scene, the geometric abstraction module first transforms it into an abstract geometric scene representation. The conditioning module generates a temporal series of abstract geometric scene representations, either based on prior routines from known OR events or based on user-generated trajectories on the geometric representation of the initial scene. The diffusion module -- built on a video-to-video diffusion backbone -- then uses the initial scene and the series of abstract geometric scene representations (video conditioning) as input to diffuse an OR event. 

\begin{figure}[!b]
    \centering
    \includegraphics[width=0.95\linewidth]{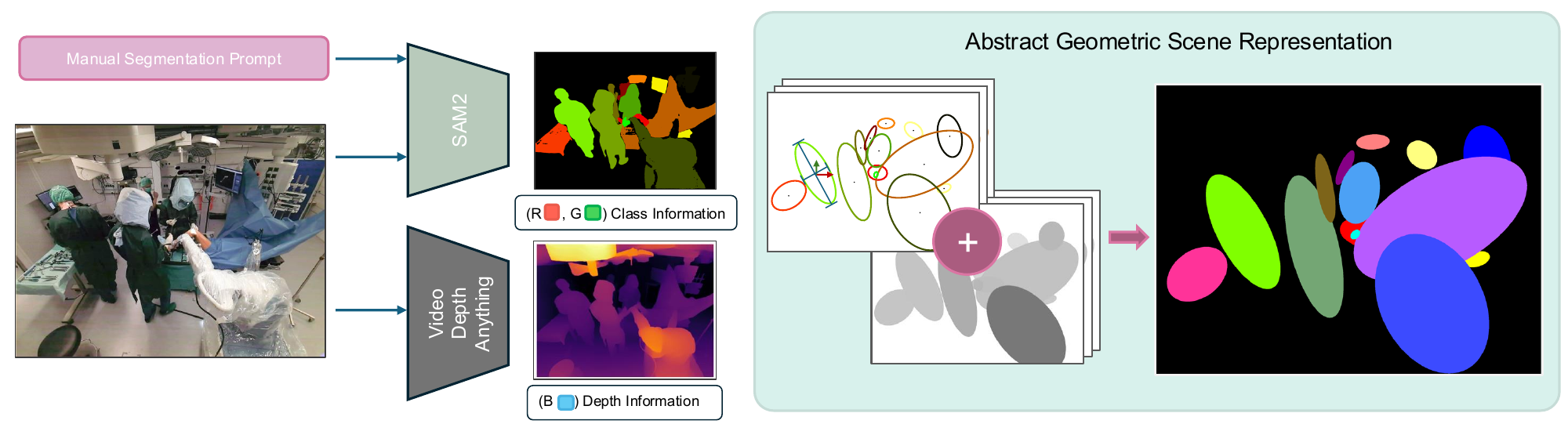}
\caption{Geometric abstraction module pipeline: Given an initial scene and segmentation point prompts, \uline{SAM2}~\cite{ravi2024sam} propagates instance segmentation masks across video. Depth information is estimated using \uline{Video Depth Anything}~\cite{chen2025video}. Each segmented instance is then approximated by an ellipsoid parameterized by its centroid position and spatial spread (height, width, rotation angle). The resulting \uline{Abstract Geometric Scene Representation} encodes class information in the red and green channels, combined with normalized relative depth in the blue channel intensity. }

\label{fig:ellipsoid_modeling}
\end{figure}


\noindent \textbf{Abstract Geometric Scene Representation}: 
The abstract geometric scene representation can be formulated as an implicit scene graph $\mathcal{G} = (\mathcal{V},\mathcal{E})$, where $\mathcal{V} = \{v^1, v^2, \ldots, v^k\}$ represents a set of $k$ nodes corresponding to OR entities (OR personnel, patient, and equipment), and edges ($\mathcal{E}$) represent the implicit spatial and temporal relationships between the nodes. Each node $v^j$ consists of geometric attributes $g^j \in \mathbb{R}^{6}$ and class information $c^j \in \mathbb{R}^{2}$. The geometric attributes encode (a) a 2-dimensional centroid position, (b) a 3-dimensional ellipsoid representation capturing spatial spread (height, width, and rotation angle), and (c) a 1-dimensional normalized relative depth value. The 2D class vector $c^j \in \mathbb{R}^2$ represents the (R, G) color channel values for each entity class, adopting the 36 semantic classes defined in the MMOR dataset~\cite{ozsoy2025mm}. Since the blue channel is reserved for depth encoding, the red and green channels encode semantic class information, providing visually distinct representations for each entity type. With implicit edges, spatial relationships such as proximity can be derived from pairwise distances between 2D centroids in normalized image space. Spatial layering (`in-front-of', `occluded-by') is captured through relative depth values. Temporally, nodes representing the same object across frames form implicit correspondences.

\uline{Rendering abstract scene representation:} Each scene representation ($\mathcal{G}$) is rendered as a 2D image at $1024 \times 768$ resolution (Fig.~\ref{fig:ellipsoid_modeling}), with nodes depicted as ellipses on a black canvas. Each ellipse is positioned at its centroid and scaled and rotated according to its spatial spread parameters (height, width, and rotation angle). Class information is encoded in the red and green color channels, providing unique colors for each object class, while normalized depth is encoded in the blue channel intensity. 

\noindent \textbf{(i) Geometric abstraction module: }
Given an OR scene, the abstract geometric scene representation is created using a semi-automated pipeline that employs out-of-the-box SAM2~\cite{ravi2024sam} and Video Depth Anything~\cite{chen2025video} models. Firstly, entities in the scene and their geometric parameters (ellipsoids' centroid position, height, width, and rotation angle) are extracted using segmentation masks generated through SAM2 using manual (inference)/groundtruth (training) prompts. The 1D normalized relative depth values are extracted using depth maps and are averaged over an instance mask. The extracted features are then used to render the abstract scene representation.

\begin{figure}[!b]
    \centering
    \includegraphics[width=0.95\linewidth]{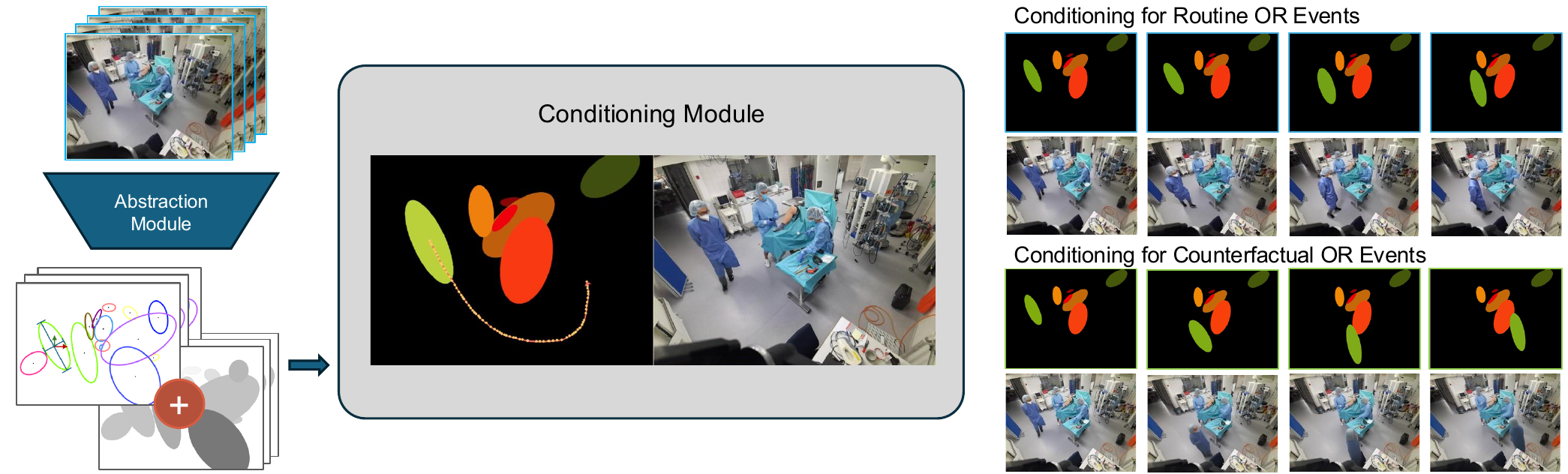}
    \caption{Interactive conditioning module for counterfactual event generation. Given an input OR video sequence, the \uline{Abstraction module} converts the scene into an abstract geometric representation. A graphical user interface enables direct manipulation of these ellipsoids through drag-and-drop operations to sketch desired trajectories. The \uline{Conditioning Module} transforms the original geometric sequence into a counterfactual event by incorporating the user-modified trajectories.}
    \label{fig:couterfactual_workflow}
\end{figure}
\noindent \textbf{(ii) Conditioning module:}
With the video generation formulated as video-to-video diffusion task, this module curates a series of abstract scene representations (corresponding to the number of frames in the diffused video) to condition the diffusion. Firstly, it employs the geometric abstraction module on all the frames of template videos of a known OR event. The resulting series of abstract representations is then used as conditioning on the initial scene to diffuse the synthetic OR event. Alternatively, the module also offers the flexibility to alter trajectories of one or more entities using user-defined trajectories, to diffuse synthetic atypical/rare/safety-critical OR events. We implement an interactive trajectory drawing tool (using OpenCV, Pygame, and Tkinter) that allows users to select ellipses from the abstract geometric representation by clicking on them, then sketch desired movement paths by drawing freehand trajectories. The tool captures waypoints along the drawn trajectory, which are interpolated across the full video sequence and applied as translational offsets to the selected ellipse's centroid position. Fig.~\ref{fig:couterfactual_workflow} illustrates an example of video conditioning using user-defined trajectories: OR-personnel walking around the instrument table, instead of moving towards the patient.

\noindent \textbf{(iii) Diffusion module:}
We employ LTX-Video~\cite{hacohen2024ltx} -- a transformer-based latent video diffusion model as the diffusion backbone model. Specifically, we fine-tuned using the In-Context LoRA (IC-LoRA) pipeline, that allows for video-video diffusion by conditioning on reference frames (abstract geometric scene representation). During fine-tuning, in addition to baseline implementation, we incorporate PatchGAN loss~\cite{isola2017image} to further improve the fidelity and realism of the synthesized video. The fine-tuned IC-LoRA weights, trained specifically for in-context conditioning, enable the model to interpret these rendered visualizations as structural guidance during generation.

\section{Experiments and Results}\label{sec:experiments_results}




\subsection{Experimental setup:}

\textbf{(i) Dataset}: We employ two public datasets: MMOR~\cite{ozsoy2025mm} and 4DOR~\cite{ozsoy20224d}. The diffusion model is trained and validated on videos from the MMOR dataset. With original videos available at 1 fps, we temporally interpolate it to $24$ fps using the LTX's keyframe interpolation feature~\cite{hacohen2024ltx}. To maintain segmentation consistency across the interpolated video, we employ SAM2 to segment OR entities based on the first-frame groundtruth annotations. Videos are processed at $1024 \times 768$ resolution, with 97 frames each. The train/test split was assigned on a video-wise basis: $338$ videos for fine-tuning the framework and $50$ videos for a detailed ablation study. For baseline in- and out-of-domain comparison against baseline conditional diffusion models, we used $6$ videos MMOR testset and $6$ videos from 4DOR.

\noindent \textbf{(ii) Training and inference.} 
\uline{Diffusion model training and inference:} The IC-LoRA-adapted video diffusion model is trained on a single NVIDIA A100 GPU for 8000 steps. We adopt the default hyperparameters from LTX's video style transfer configuration: LoRA rank and alpha of 128, learning rate of $2 \times 10^{-4}$, AdamW optimizer, and bfloat16 mixed precision training. Inference is performed using 50 denoising steps with a guidance scale of 3.5. During training, first-frame conditioning is provided in 20\% of cases to encourage both conditional and unconditional generation capabilities. At inference time, first-frame conditioning is performed exclusively, providing the initial frame of the target video alongside the complete rendered abstract scene representations. 

\noindent \textbf{(iii) Evaluation Metrics :} \uline{(a) With Groundtruth videos:} We use FVD, SSIM, and PSNR, and LPIPS metrics. SSIM and PSNR are used to quantify the average video quality and degradation of the generated videos against the groundtruth videos. FVD summarizes set‑level spatio‑temporal realism. Additionally, to evaluate structural accuracy and alignement with abstract conditioning, we use bounding box IoU (BB IoU) and segmentation mask IoU (Seg IoU). These metrics quantify controllability by measuring the spatial alignment between the conditioned ellipsoid trajectories and the corresponding entity positions in the generated video. We prompt each instance in the groundtruth initial frame for generating segmentation masks using SAM2 and track them across both the real and generated video sequences. By comparing the resulting segmentation masks and bounding boxes between real and generated sequences, we measure spatial alignment and component localization accuracy.
\uline{(b) Without groundtruth videos:} we employ DOVER~\cite{wu2023exploring} and Inception Score~\cite{salimans2016improved} to quantify performance. \uline{(c) Downstream near-miss detection task:} We prioritize recall (sensitivity) over accuracy, as missing a true violation (false negative) poses greater clinical risk than a false alarm (false positive), which simply prompts staff verification.

\subsection{Results}\begin{table}[!t]
    \centering
    \caption{Comparison of our framework against out-of-the-box baseline models on in- and out-of-domain testsets. \uline{WAN~\cite{wan2025wan} \& LTX-base (LTX\textsubscript{b})~\cite{hacohen2024ltx}}: Text-conditioned generation using VLM descriptions of the groundtruth scene. \uline{SVD~\cite{blattmann2023stable}}: Image-to-video generation with low dynamic motion setting. \uline{Ours}: Our proposed framework.}
        \begin{tabular}{lcccc|cccc}
        \toprule
        \textbf{Method} & FVD$\downarrow$ & SSIM$\uparrow$ & PSNR$\uparrow$ & LPIPS$\downarrow$  & FVD$\downarrow$ & SSIM$\uparrow$ & PSNR$\uparrow$ & LPIPS$\downarrow$ \\
        \midrule
        \multicolumn{5}{c|}{\textbf{MMOR (In-Domain)}} & \multicolumn{4}{c}{\textbf{4DOR (Out-of-Domain)}} \\
        \midrule
        WAN~\cite{wan2025wan} & 1190.57 & 0.78 & 18.95 & 0.20 & 699.78 & 0.86 & 21.72 & 0.13 \\
        SVD~\cite{blattmann2023stable} & 5021.19 & 0.68 & 17.91 & 0.24 & 3790.73 & 0.74 & 19.86 & 0.16 \\
        LTX\textsubscript{b}~\cite{hacohen2024ltx} & 2439.33 & 0.47 & 12.88 & 0.58 & 1135.26 & 0.46 & 13.10 & 0.58 \\
        Ours & \textbf{689.88} & \textbf{0.86} & \textbf{23.21} & \textbf{0.13} & \textbf{265.25} & \textbf{0.90} & \textbf{25.87} & \textbf{0.07} \\
        \bottomrule
        \end{tabular}
\label{tab:video_quality_comparison}
\end{table}

\begin{figure}[!t]
    \centering
    \includegraphics[width=1.0\linewidth]{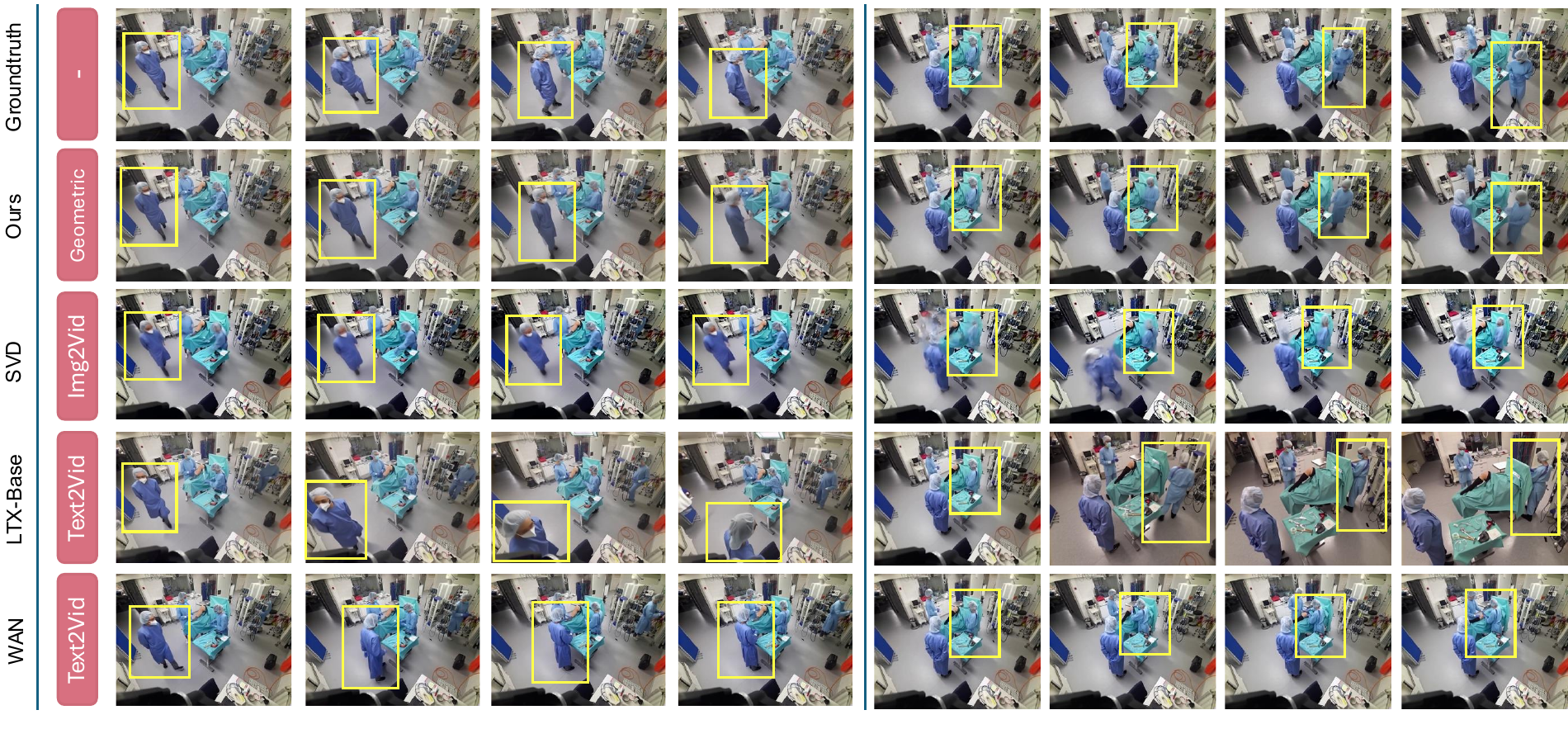}
    \caption{Qualitative comparison of video synthesis methods on out-of-domain (4DOR) dataset. \uline{Groundtruth}: Original video frames to reconstruct. \uline{WAN~\cite{wan2025wan} \& LTX-Base}: Text-conditioned generation using VLM descriptions of the groundtruth scene. \uline{SVD}: Image-to-video generation with low dynamic motion setting. \uline{Ours}: Our proposed video synthesis using abstract geometric representation. }

    \label{fig:results1}
\end{figure}

\textbf{(i) Baseline comparison:}
Firstly, we benchmark our abstract geometric conditioned OR diffusion framework on in- and out-of-domain testsets against out-of-the-box baseline models: (i) SVD~\cite{blattmann2023stable} that performs image to video diffusion, and (ii) WAN~\cite{wan2025wan} and LTX-base~\cite{hacohen2024ltx} that condition video via text-prompt. Quantitatively (Table.~\ref{tab:video_quality_comparison}), our framework -- fine-tuned on a small trainset (338 videos; 97 frames each) -- performs well on both in-domain and out-of-domain testsets. Fig.~\ref{fig:results1} shows the qualitative performance of our framework against baseline models on out-of-domain videos. These results demonstrate the effectiveness of our abstract geometry conditioning OR video generation framework in conditioning the OR video synthesis at every frame, for each entity.

\begin{figure}[!t]
    \centering
    \includegraphics[width=0.98\linewidth]{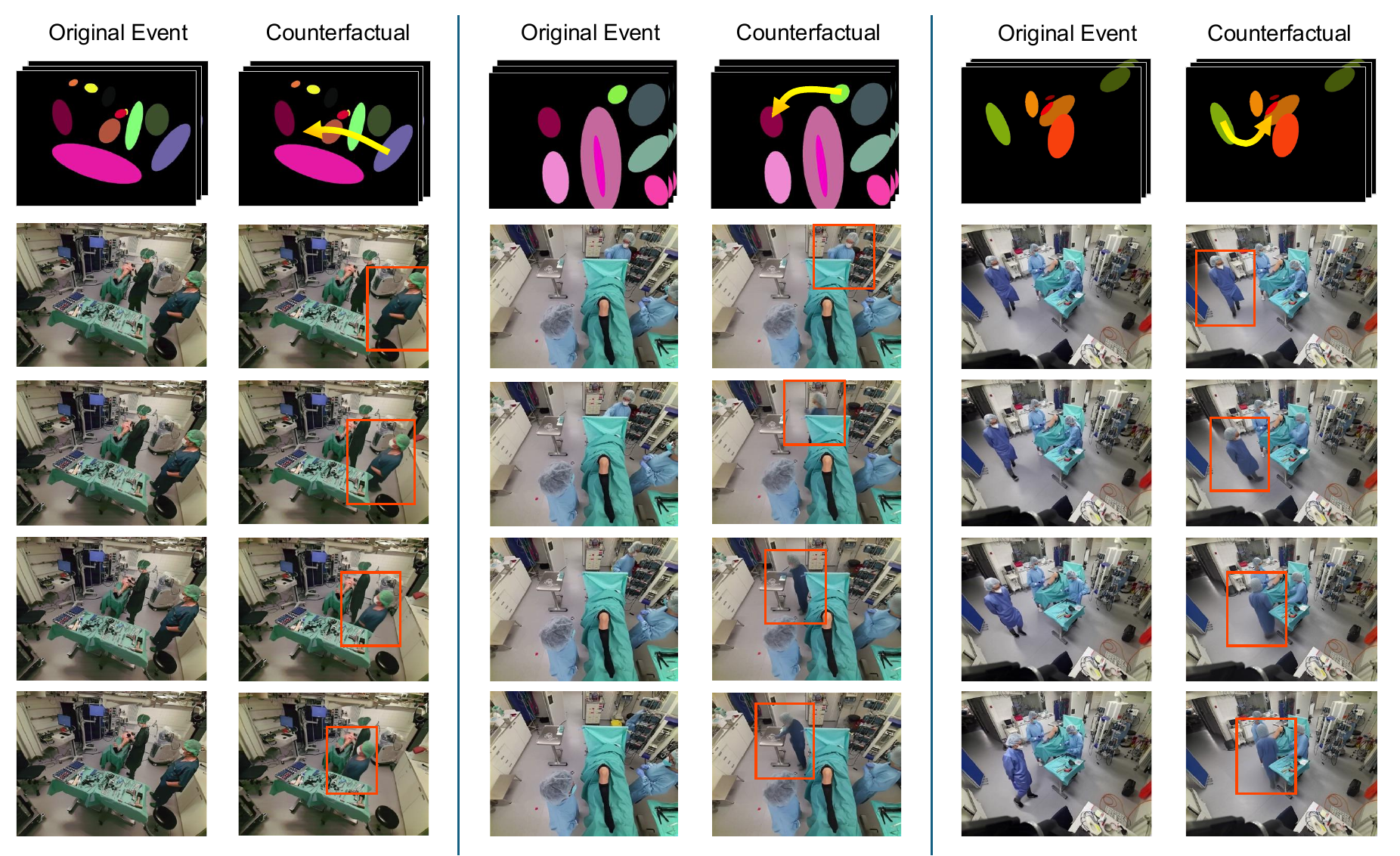}
    \caption{Controllable synthesis of safety-critical, interactions, and alternate OR events. Each column pair shows a routine OR event (left) with its abstract geometric representation (top), and a counterfactual event (right) generated by providing a trajectory for geometric conditioning. \uline{Left pair (safety-critical event):} A non-sterile assistant approaches the sterile instrument table. \uline{Middle pair (Interaction):} Personnel walking toward and reaching for interaction with the table. \uline{Right pair (Alternate event):} Modified trajectory where personnel walks directly toward the patient bed instead of the original path around the room.
}
    \label{fig:results2}
\end{figure}

\noindent \textbf{(ii) Synthesizing rare/atypical/satefy-critical events:} To demonstrate the framework's flexibility in controlled videos synthesis of rare/atypical/safety-critical events, which would otherwise be difficult to generate without straining the workforce or risking patient harm, we qualitatively (Fig.~\ref{fig:results2}) and quantitatively (Table.~\ref{tab:quality_comparison}) assess its performance. Quantitatively, we show that our framework performs better than out-of-the-box DragNUWA~\cite{yin2023dragnuwa} -- a baseline model that also conditions video generation through user-defined sketch. We show that, by using an interactive conditioning tool, where ellipsoids (geometric representation of OR entities) can be manipulated and dragged to generate new events, the framework can diffuse counterfactual OR events. Qualitatively, we observe that, while our framework allows for explicit spatial conditioning of entity trajectories, in some cases, it has also implicitly learned the interactions between entities in the training distributions based on spatial proximity. For instance, when an OR personnel is conditioned to move towards the instrument table, the framework diffuses an OR event video, where the person is seen interacting with the table.

\begin{table}[!t]
\centering
\begin{minipage}{0.40\textwidth}
\centering
\caption{Quantitative assessment of our framework's ability in curating synthetic rare/atypical/safety-critical OR events.}
\label{tab:quality_comparison}
\begin{tabular}{lcc}
\toprule
Method & DOVER$\uparrow$ & Inception Score$\uparrow$ \\
\midrule
DragNUWA~\cite{yin2023dragnuwa} & 0.31 & $1.04 \pm 0.05$ \\
Ours & \textbf{0.52} & $1.03 \pm 0.01$ \\
\bottomrule
\end{tabular}
\end{minipage}
\begin{minipage}{0.40\textwidth}
\centering
\caption{Detecting near safety-critical events (near misses of sterile-field violation) using models trained on synthetic data.}
\label{tab:application_results}
\begin{tabular}{l|cc}
\toprule
Method & Accuracy$\uparrow$ & RECALL$\uparrow$ \\
\midrule
ResNet34~\cite{he2016deep} & 64.91 & 50.65 \\
ViT-B/16~\cite{dosovitskiy2020image} & \textbf{67.54} & \textbf{70.13} \\
\bottomrule
\end{tabular}
\end{minipage}
\end{table}

\begin{figure}[!    t]
    \centering
    \includegraphics[width=0.95\linewidth]{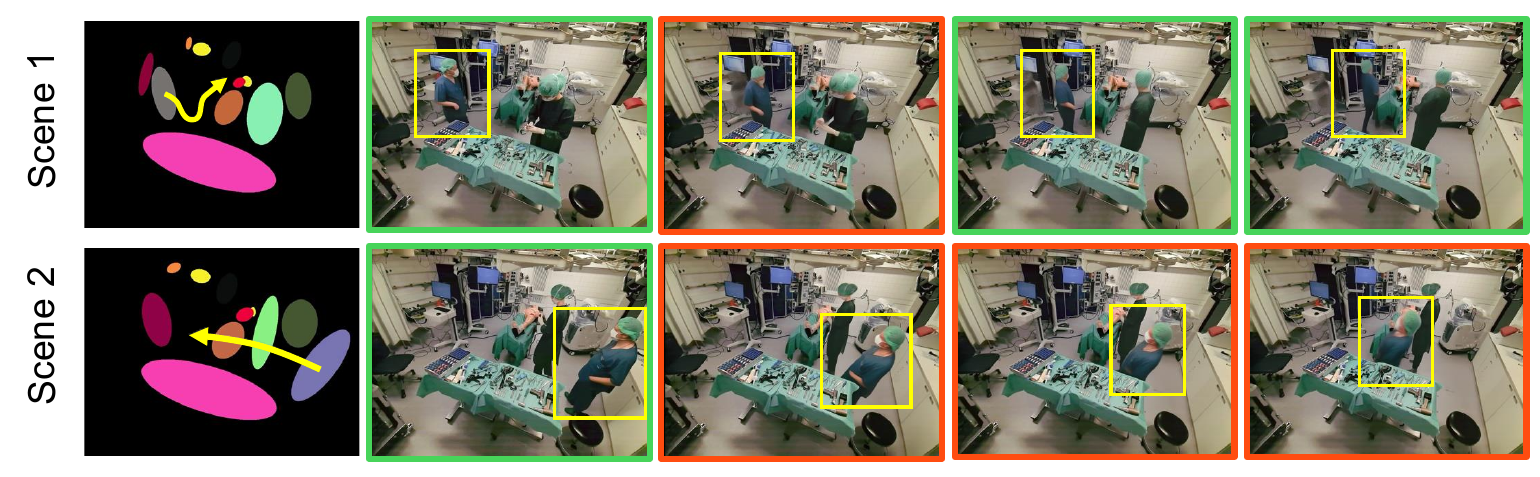}
    \caption{Left: abstract geometric conditioning. Right: synthesized video frames. Positive (red) and negative (green) training samples for near safety-critical event (near misses in sterile-field violation) detection from counterfactual synthetic data generated using our framework. Two scenes demonstrate near-miss progressions: \uline{Scene 1} shows non-sterile personnel approaching then retreating from the instrument table. \uline{Scene 2} shows personnel passing close to the instrument table. }

    \label{fig:Figure6}
\end{figure}

\noindent \textbf{(iii) Developing OR ambient intelligence model for detecting near safety-critical events from synthetic data:}
Considering sterile-field violations can potentially compromise patient outcomes, we define near misses in sterile-field violations as scenarios in which non-sterile personnel approach the sterile field without making contact, and treat these instances as near safety-critical events.
Using our trained framework, we curate a synthetic dataset to train and validate AI models for detecting near-misses of sterile-field violations. Using $20$ of the $50$ MMOR testset videos, we curated $87$ synthetic videos, depicting positive and negative samples for near-misses of sterile-filed violations. Image frames from $68$ of these synthetic videos were used to train the model. Frames from the remaining videos were used for model validation. The synthetic dataset comprises 678 training frames (252 positive, 426 negative) and 228 validation frames (77 positive, 151 negative). Positive samples represent frames where non-sterile personnel are in close proximity to the sterile field, while negative samples represents a OR scene where non-sterile personnel maintain a safe distance from the sterile field. The near-miss detection model is a per-frame image classifier that operates on individual frames without trajectory history, detecting near-misses based on spatial proximity within each frame. Fig.~\ref{fig:Figure6} shows positive and negative samples of these frames generated from conditioning the trajectories of entities using the framework. Table~\ref{tab:application_results} summarizes the performance of classifiers trained and validated on these synthetic samples for detecting near-misses of sterile-field violations.

\begin{table}[!t]
\centering
\caption{Ablation study of our framework with and without base (diffusion backbone and baseline fine-tuning), Seg (segmentation mask-based conditioning), E (ellipsoids-based conditioning), D (depth encoded ellpsoids) and $L_g$ (PatchGAN loss integrated for finetuning).}
\begin{tabular}{ccccc|cccccc}
\toprule
\multicolumn{5}{c|}{Method} & \multirow{2}{*}{FVD$\downarrow$} & \multirow{2}{*}{SSIM$\uparrow$} & \multirow{2}{*}{PSNR$\uparrow$} & \multirow{2}{*}{LPIPS$\downarrow$} & \multirow{2}{*}{BB IoU$\uparrow$} & \multirow{2}{*}{Seg IoU$\uparrow$} \\
Base  & Seg & E & D & $L_{g}$ &                      &                       &                       &                        &                         &                          \\
\midrule
\checkmark  & \checkmark  & & &  & \textbf{347.88} & \textbf{0.88} & \textbf{25.34} & \textbf{0.09} & \textbf{0.96} & \textbf{0.95} \\
\checkmark  &  & \checkmark& &  & 518.50 & 0.86 & 23.65 & 0.12 & 0.93 & 0.90\\
\checkmark  &  &  \checkmark & \checkmark &  & 532.05 & 0.86 & 23.74 & 0.12 & 0.93 & 0.90 \\
\checkmark  & & \checkmark & \checkmark & \checkmark & 487.20 & 0.88 & 24.71 & 0.11 & 0.93 & 0.91 \\
\bottomrule
\end{tabular}
\label{tab:quantitative}
\end{table}

\noindent \textbf{(iv) Ablation Study:} We perform extensive ablation study using all $50$ MMOR testset videos to validate the key components of our framework. Table~\ref{tab:quantitative} compares three conditioning approaches: direct segmentation maps, ellipse rendering without depth, and our proposed ellipse rendering with depth encoding and adversarial training. 
While segmentation map conditioning achieves superior reconstruction metrics (FVD: 347.88), this comes at the cost of controllability as segmentation masks are not easy to manipulate (such as fine-grained conditioning the limbs) or move around. Our ellipse-based representation maintains strong absolute performance (SSIM $>$ 0.88, and BBox IoU $>$ 0.93) while enabling flexibility in conditioning and scene composition. Adding PatchGan loss ($L_g$) further enhances performance to an FVD of 487.20 and segmentation IoU of 0.91.

\section{Discussion and Conclusion}
In this work, we introduced an OR video diffusion framework that reformulates a video-to-video diffusion task as an OR event diffusion model conditioned on abstract geometry scene representation. By abstracting the input scene and routine OR events into a visualizable geometric representation, and coupling it with an interactive conditioning module, our framework offers a flexible and controlled diffusion of OR events. This unlocks synthetic video generation of routine, atypical, rare, and safety‑critical OR events, at scale, that otherwise are difficult/near-impossible to curate due to strain on the workforce and risk to patient outcome. We demonstrate that our framework -- fine-tuned on a small public dataset of 338 videos -- outperforms out-of-the-box baseline models both quantitatively and qualitatively on small in-domain and out-of-domain test sets. We also showcase our framework's controllability in generating atypical/rare/safety-critical OR events using abstract geometric scene conditioning. Additionally, we show our framework's potential in generating synthetic data towards training AI models for detecting near safety-critical events -- near misses of sterile-field violation. 

With this work serving as a groundwork toward scalable data curation for developing OR ambient intelligence for OR workflow analysis, key limitations exist that need to be progressively addressed.
\textit{(i) Conditioning and controllability tradeoff:} We selected ellipsoids as geometric primitives to enable intuitive drag-and-drop trajectory control while being robustly extractable from segmentation masks, unlike articulated pose representations that would require complex interfaces and are prone to failure in cluttered OR scenes. This design choice provides sufficient granularity for spatial-conditioning to enforce proximity/movement of OR personnel near OR devices, but is limited in enforcing explicit fine-grained articulation control (e.g., arm extension of OR personnel when reaching for instruments). In the current framework, the generative model implicitly learns articulation and interaction priors from the training data distribution, guided by spatial‑proximity conditioning.
\textit{(ii) Generalization and robustness:} Our framework demonstrated generalizability to out-of-domain dataset (4DOR) for routine event synthesis. However, challenges remain in generating synthetic videos for significantly different environments due to variations in sterile attire colors, equipment configurations, and surgeries (e.g., open surgery, emergency trauma) not represented in the MMOR training data.
\textit{(iii) Clinical validation and downstream utility:} While clinical collaborators were consulted throughout the development stages, with this study being groundwork towards developing scalable synthetic data generation for OR ambient intelligence, formal domain‑expert evaluation (e.g., structured ratings by independent surgeons) is beyond the scope of this work. The downstream ambient AI model -- near critical-event detection model -- trained and validated on synthetic data, serves as a proof-of-concept. Comprehensive evaluation on real OR images and the impact of ambient intelligence in enhancing hospital performance -- clinical and financial -- remains to be studied.

Future work includes: (i) further improving the video‑diffusion model's performance. Although integrating a PatchGAN loss during fine‑tuning has improved fidelity, the resolution and clarity of moving entities degrade as the video progresses and deviates from the groundtruth event; architectural modifications to enhance temporal fidelity and consistency are a promising direction. (ii) Introducing scalable, intuitive conditioning for explicit fine‑grained articulation conditioning (e.g., OR personnel reaching for instruments), extending beyond the current framework's explicit spatial (trajectory) conditioning. (iii) Reducing reliance on manual SAM2 prompts at inference by developing a scalable data‑curation pipeline that automates geometric abstraction via zero‑shot entity detection, enabling large‑scale processing of public OR datasets. (iv) Comprehensive validation, including clinical validation, on framework generalizability to real and diverse OR environments.


\backmatter

\section*{Declarations}

\noindent \textbf{Funding:} This work was funded by the National Science Foundation, under Grant No. 2239077. The content is solely the responsibility of the authors and does not necessarily represent the official views of the National Science Foundation.

\noindent \textbf{Competing interests:} The authors have no competing interests.

\noindent \textbf{Ethics approval:} This is a computational study involving no human participants or animals and is based on publicly available datasets. No ethical approval was required.

\noindent \textbf{Informed consent:} Not applicable.

\noindent \textbf{Author contributions:} The first two authors contributed equally to this work.

\bibliography{references}

\end{document}